\documentclass[sigconf]{acmart}
\renewcommand\footnotetextcopyrightpermission[1]{} 

\AtBeginDocument{%
  }

\usepackage{graphicx}
\usepackage{subcaption}
\usepackage{multirow}
\usepackage{amsmath}
\usepackage[linesnumbered,ruled]{algorithm2e}

\usepackage{siunitx}

\begin{document}

\title{Investigating the Robustness of Counterfactual Learning to Rank Models: A Reproducibility Study}

\author{Zechun Niu}
\affiliation{%
  \department{Gaoling School of Artificial Intelligence}
  \institution{Renmin University of China}
  \city{Beijing}
  \country{China}
}
\email{niuzechun@ruc.edu.cn}

\author{Zhilin Zhang}
\affiliation{%
  \department{School of Statistics}
  \institution{Renmin University of China}
  \city{Beijing}
  \country{China}
}
\email{2023201838@ruc.edu.cn}

\author{Jiaxin Mao}
\authornote{Corresponding author.}
\affiliation{%
  \department{Gaoling School of Artificial Intelligence}
  \institution{Renmin University of China}
  \city{Beijing}
  \country{China}
}
\email{maojiaxin@gmail.com}

\author{Qingyao Ai}
\affiliation{%
  \department{Department of Computer Science and Technology}
  \institution{Tsinghua University}
  \city{Beijing}
  \country{China}
}
\email{aiqy@tsinghua.edu.cn}

\author{Ji-Rong Wen}
\affiliation{%
  \department{Gaoling School of Artificial Intelligence}
  \institution{Renmin University of China}
  \city{Beijing}
  \country{China}
}
\email{jirong.wen@gmail.com}


\begin{abstract}
  Counterfactual learning to rank (CLTR) has attracted extensive attention in the IR community for its ability to leverage massive logged user interaction data to train ranking models. While the CLTR models can be theoretically unbiased when the user behavior assumption is correct and the propensity estimation is accurate, their effectiveness is usually empirically evaluated via simulation-based experiments due to a lack of widely available, large-scale, real click logs. However, many previous simulation-based experiments are somewhat limited because they may have one or more of the following deficiencies: 1) using a weak production ranker to generate initial ranked lists, 2) relying on a simplified user simulation model to simulate user clicks, and 3) generating a fixed number of synthetic click logs. As a result, the robustness of CLTR models in complex and diverse situations is largely unknown and needs further investigation. 
  
  To address this problem, in this paper, we aim to investigate the robustness of existing CLTR models in a reproducibility study with extensive simulation-based experiments that (1) use production rankers with different ranking performance, (2) leverage multiple user simulation models with different user behavior assumptions, and (3) generate different numbers of synthetic sessions\footnote{In this paper, a ``session'' refers to the set of events between a user's issuing a query and abandoning the search result page.} for the training queries. We find that the IPS-DCM, DLA-PBM, and UPE models show better robustness under various simulation settings than other CLTR models. Moreover, existing CLTR models often fail to outperform naive click baselines when the production ranker is strong and the number of training sessions is limited, indicating a pressing need for new CLTR algorithms tailored to these conditions.

\end{abstract}

\ccsdesc[500]{Information systems~Learning to rank}

\keywords{Counterfactual Learning to Rank; Robustness; Simulation-based Experiment}

\maketitle

\section{Introduction}
\label{section_1}
Users' interaction with search systems (e.g., clicks, dwell time) is a valuable resource because it not only reflects users' relevance preferences for the documents but also is easy to collect at scale. By logging user interaction data, it is possible to estimate the user-centric utility of the documents and train powerful but data-hungry ranking models (e.g., neural ranking models \cite{guo2016deep, mitra2017learning}). However, user interaction data is noisy and biased \cite{joachims2005accurately, joachims2007evaluating, o2006modeling, yue2010beyond}. For example, users are inclined to examine the documents at higher ranks, which may result in position bias in click logs \cite{joachims2005accurately}. As a result, directly training a ranking model to optimize click data can lead to poor performance. Therefore, many researchers have proposed various counterfactual learning to rank (CLTR) models \cite{wang2018regression-EM, vardasbi2020affine} to train an unbiased ranking model with biased user interaction data.

A counterfactual learning to rank model is usually constructed based on a specific user behavior model, such as position-based click model (PBM) \cite{craswell2008experimental}, which assumes that the probability of users examining a document is only related to its position in the ranked list. Under such assumptions, it estimates the probability that a document is examined (called ``propensity'' from a counterfactual view) and removes position bias from the user interaction data through weighing each click with its inverse propensity weight (IPW) \cite{ipw}. Theoretically, Joachims et al.~\cite{joachims2017ips} proved that the counterfactual learning to rank framework is unbiased when the propensity estimation is accurate and non-zero for every relevant document.

Because these two ideal assumptions above may not always hold in real application scenarios, there is no guarantee that theoretically principled CLTR models could be effective and robust in practice. Due to a lack of large-scale real user behavior logs, researchers often use simulation-based experiments as an alternative method to evaluate CLTR models. However, previous simulation-based experiments \cite{joachims2017ips, ai2018dla, vardasbi2020cm-ips, chen2021iobm, wang2021prs, Luo2024UPE} were not comprehensive enough and had the following defects. Firstly, most of them \cite{joachims2017ips, ai2018dla, vardasbi2020cm-ips, chen2021iobm, wang2021prs} used one percent of labeled training data to train a Ranking SVM model \cite{joachims2006SVMRank} to generate initial ranked lists. As a result, these simulation-based experiments are limited to a production ranker with a relatively poor ranking performance, which casts doubt on whether the CLTR models work well for a more sophisticated and effective real online search system. Secondly, many of them \cite{joachims2017ips, ai2018dla, wang2021prs, Luo2024UPE} only used the PBM user simulation model to simulate user clicks. However, the user behavior assumption of PBM is rather simple and cannot simulate various user behavior patterns in the real world. Thirdly, some of them \cite{ai2018dla, chen2021iobm, Luo2024UPE} generated a sufficient number of synthetic click logs and did not evaluate how the CLTR models would be affected by variance when dealing with sparse click data. Therefore, we argue that the popular simulation-based experiments cannot fully verify the robustness of CLTR models in various situations and thus cannot guarantee their performance in complex and diverse real-world scenarios \cite{Baidu}. 

In response to these shortcomings, our study aims to further investigate the robustness of CLTR models and address the following research questions (RQ):
\begin{itemize} \item {RQ1}: How robust are the CLTR models under various simulation settings? 
\item {RQ2}: How does the production ranker affect the CLTR models?
\item {RQ3}: How does the user simulation model affect the CLTR models?
\item {RQ4}: How does the size of synthetic click logs affect the CLTR models?
\end{itemize}

To answer the above research questions, we conduct a reproducibility study of existing CLTR models and extend the mainstream simulation-based experiments in the following directions:
\begin{itemize}
\item {Using more effective production rankers:} Previous simulation-based experiments \cite{joachims2017ips, ai2018dla, vardasbi2020cm-ips, chen2021iobm, wang2021prs} typically used one percent of labeled training data to train a Ranking SVM model as the production ranker without trying a production ranker with better ranking performance. To bridge the gap, we use different percentages of labeled training data to train Deep Neural Networks (DNN) with different ranking performance as the production rankers.
\item {Incorporating different types of user simulation models:} Previous simulation-based experiments \cite{joachims2017ips, ai2018dla, vardasbi2020cm-ips, chen2021iobm, wang2021prs} usually used PBM as the only user simulation model, whose assumptions are too simple to simulate the various user behavior patterns in various real-world scenarios. Therefore, in addition to PBM, we also utilize two different user simulation models: one is the dependent click model (DCM), which is one of the popular cascade-based models; the other is the comparison-based click model (CBCM) \cite{zhang2021cbcm}, which assumes that users would compare adjacent documents to make click-through decisions.
\item {Varying the size of synthetic click logs:} Some previous simulation-based experiments \cite{ai2018dla, chen2021iobm, Luo2024UPE} generated a fixed number of synthetic click logs and did not evaluate the performance of the CLTR models when facing different scales of click data. To compensate for this shortcoming, we generate different numbers of synthetic click logs to train the CLTR models.
\end{itemize}

The contributions of this paper are summarized as follows:
\begin{itemize}
\item We are the first to systematically extend the simulation-based experiments, tailored to the evaluation of counterfactual learning to rank (CLTR) models.
\item We propose a new metric to investigate and compare the robustness of a wide range of CLTR models under a variety of simulation settings.
\item We explore the effect of production rankers, user simulation models, and sizes of synthetic click logs on the CLTR models, which may explain when and why their ranking performance is suboptimal in certain situations.
\item We open-source a toolkit for reproducing the training and evaluation of CLTR models, which could be used as a more comprehensive benchmark for future research on unbiased learning to rank.\footnote{\href{https://github.com/Diligentspring/Investigating-the-Robustness-of-Counterfactual-Learning-to-Rank-Models}{https://github.com/Diligentspring/Investigating-the-Robustness-of-Counterfactual-Learning-to-Rank-Models}}
\end{itemize}

\section{Counterfactual Learning to Rank}
\label{sec:preliminaries}
In this section, we introduce some classic counterfactual learning to rank (CLTR) models that aim to learn unbiased ranking models with biased historical click logs. CLTR models follow an examination hypothesis (EH) \cite{richardson2007ExminationHypothesis}, which assumes that the clicks $c$ can be broken down into examination $o$ and relevance judgment $r$:
\begin{equation}
\label{eq: EH}
    P(c_x^{q} = 1) = P(o_x^{q} = 1) \times P(r_x^{q} = 1),
\end{equation}
where $x$ and $q$ denote a document and a query, respectively. Therefore, the core of CLTR is to reveal the true relevance from the clicks by eliminating the impact of examination.

\subsection{Inverse Propensity Scoring (IPS)}
Wang et al. \cite{wang2016ips} and Joachims et al. \cite{joachims2017ips} presented a counterfactual inference framework that provides the theoretical basis for unbiased learning to rank via Empirical Risk Minimization. Taking the examination probability as ``propensity'', they leveraged the inverse propensity scoring (IPS) \cite{ipw} method to estimate the local empirical risk $R(S|q)$ of any new ranker $S$ on arbitrary query $q$: 
\begin{equation}
\label{eq_ips}
    \hat{R}_{IPS}(S|q) =  \sum\limits_{x \in \pi_q, c_x^{q}=1} \frac{\Delta(x|\pi_q)}{P(o_x^{q}=1|\pi_q)} \quad ,
\end{equation}
where $\pi_q$ denotes the ranked list returned by $S$, and $\Delta(x|\pi_q)$ denotes a function that computes the contribution from the document $x$ to the loss function, which is usually related to the ranking metric. For example, if we use the Average Relevance Position (ARP) metric, then $\Delta(x|\pi_q) = {\rm rank}(x|\pi_q)$, where ${\rm rank}(x|\pi_q)$ denotes the rank of $x$ in $\pi_q$. Joachims et al. \cite{joachims2017ips} theoretically proved that in the ideal case where every relevant document has a non-zero probability to be examined and the correct propensity estimation can be obtained, the IPS method yields an unbiased estimate of the ground truth empirical risk $R(S)$:
\begin{equation}
    \mathbb{E}(\hat{R}_{IPS}(S)) = {R}(S)
\end{equation}

Based on the PBM \cite{craswell2008experimental} user behavior model, Wang et al. \cite{wang2016ips} and Joachims et al. \cite{joachims2017ips} proposed the IPS-PBM model:
\begin{equation}
\label{eq_ips_pbm}
    \hat{R}_{IPS-PBM}(S|q) =  \sum\limits_{x \in \pi_q, c_x^{q}=1} \frac{\Delta(x|\pi_q)}{P(o_x^{q}=1|{\rm rank}(x|\pi_q))} \quad ,
\end{equation}
where $P(o_x^{q}=1|{\rm rank}(x|\pi_q))$ is obtained through a online result randomization experiment. In addition, Vardasbi et al. \cite{vardasbi2020cm-ips} proposed a cascade model-based inverse propensity scoring (CM-IPS) method, which extends the IPS method to cascade scenarios by estimating propensities based on cascade user behavior models.

\subsection{Dual Learning Algorithm (DLA)}
A drawback of the IPS method \cite{wang2016ips, joachims2017ips} is that a separate result randomization experiment is needed to obtain accurate propensity estimates, which can negatively affect users' search experience. In response to this shortcoming, Ai et al. \cite{ai2018dla} proposed a dual learning algorithm (DLA) to jointly learn an unbiased ranking model and an unbiased propensity model. The key observation of DLA is that the problem of estimating a propensity model from click data is a dual problem of counterfactual learning to rank. Thus, similar to Eq.~(\ref{eq_ips}), the local empirical risk of the propensity model $E$ can be estimated by:
\begin{equation}
    \hat{R}_{IPS}(E| q) = \sum\limits_{x \in \pi_q, c_x^{q}=1} \frac{\Delta(x|\pi_q)}{P(r_x^{q}=1|\pi_q)} 
\end{equation}
Therefore, the empirical risk of the ranking model $S$ and propensity model $E$ can be estimated using each other's outputs, which enables us to jointly update the parameters of these two models with stochastic gradient descent (SGD).

\subsection{Propensity Ratio Score (PRS)}
Moreover, Wang et al. \cite{wang2021prs} found that the IPS method \cite{wang2016ips, joachims2017ips} ignores the bias caused by implicitly treating non-clicked documents as irrelevant when optimizing the ranking model. Although the empirical risk estimated by Eq.~(\ref{eq_ips}) is unbiased, a direct optimization of Eq.~(\ref{eq_ips}) is intractable due to the non-continuous nature of most ranking metrics. Thus, approximations are necessary for optimization \cite{agarwal2019general}, which however introduces a gap from the estimated metric to the induced loss. For example, Joachims et al. \cite{joachims2017ips} compared each clicked document to all other documents when training the Propensity SVM-Rank model, which contains loss on all ``relevant-relevant'' pairs and would distort the optimization of the ranking model. To fill the gap, Wang et al. \cite{wang2021prs} proposed propensity ratio scoring (PRS) to reweigh the pairwise losses:
\begin{equation}
\label{eq:prs}
    \Delta_{PRS}(S|q) = \sum\limits_{x_i:c_q(x_i=1)}\sum\limits_{x_j:c_q(x_j=0)} 
    \delta(x_i, x_j|\pi_q) \cdot \frac{P(o_{x_j}^{q}=1|\pi_q)}{P(o_{x_i}^{q}=1|\pi_q)}
\end{equation}
where $\delta(x_i, x_j|\pi_q)$ is the pairwise loss defined on two different documents $x_i$ and $x_j$ in $\pi_q$.

\section{Extending Simulation-based Experiments}
\label{section_methodology}
\begin{figure}[t]
  \centering
  \includegraphics[width=0.95\linewidth,trim=50 70 75 10, clip]{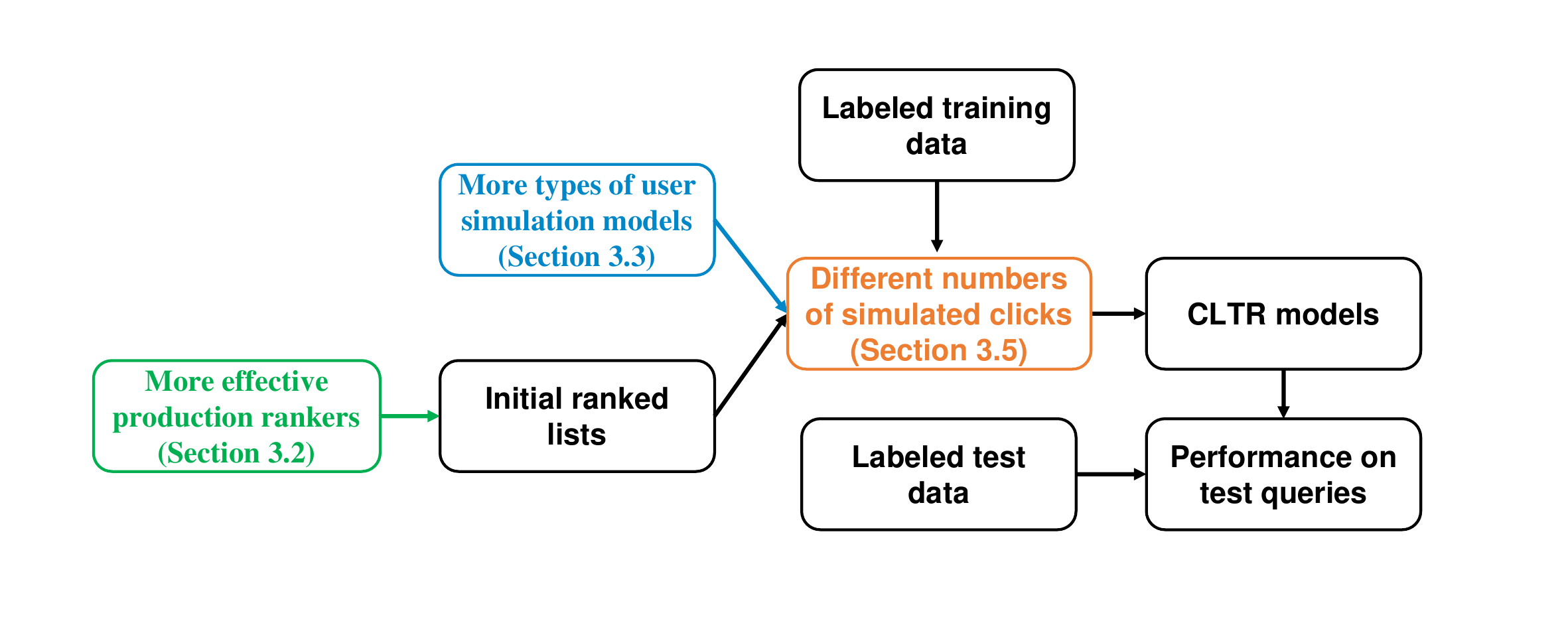}
  \vspace{-3mm}
  \caption{Overall framework of our simulation-based experiment. The components different from previous simulation-based experiments are highlighted in different colors.}
  \label{fig:simulation_framework}
  \Description{This figure shows the framework of our simulation-based experiment.}
  \vspace{-4mm}
\end{figure}

In this section, we describe how we extend the mainstream simulation-based experiments to answer the research questions presented in Section \ref{section_1}. The overall framework of our simulation-based experiment is shown in Figure~\ref{fig:simulation_framework}, where the components different from previous simulation-based experiments are highlighted in different colors. First, instead of using Ranking-SVM trained with 1\% labeled training data as the production ranker, we used deep neural networks (DNN) trained with 1\% and 20\% labeled training data as our production rankers which have more decent ranking performance. Second, we leveraged multiple user simulation models, not just PBM, to sample clicks based on the ground truth relevance labels. Third, we trained various CLTR models with different numbers of synthetic click logs. Finally, we evaluated their performance on test queries with human annotations. 

\subsection{Dataset}
We conducted a series of simulation-based experiments on two benchmark learning to rank datasets. 

\textbf{Yahoo! LETOR \cite{Yahoo!}.}\footnote{\href{https://webscope.sandbox.yahoo.com/}{https://webscope.sandbox.yahoo.com/}} It is one of the most widely used benchmarks in the field of CLTR. It contains 29,336 queries with 710k documents, and each query-document pair has 700 features and 5-level relevance labels (0-4). 

\textbf{MSLR-WEB10K \cite{MSLR}.}\footnote{\href{http://research.microsoft.com/en-us/projects/mslr/}{http://research.microsoft.com/en-us/projects/mslr/}}  It contains 10,000 queries with around 1,200k
documents. Each query-document pair has 136 features and 5-level relevance labels (0-4).

We used the set 1 of Yahoo! LETOR and Fold1 of MSLR-WEB10K, and followed the given data split of training, validation, and testing of the datasets. Following Ai et al. \cite{ai2018dla}, we also removed the queries without relevant documents and those with less than two candidate documents.

\subsection{Production Ranker}
\begin{figure}[t]
  \centering
  \vspace{-1mm}
  \includegraphics[width=0.6\linewidth, trim=10 0 40 20, clip]{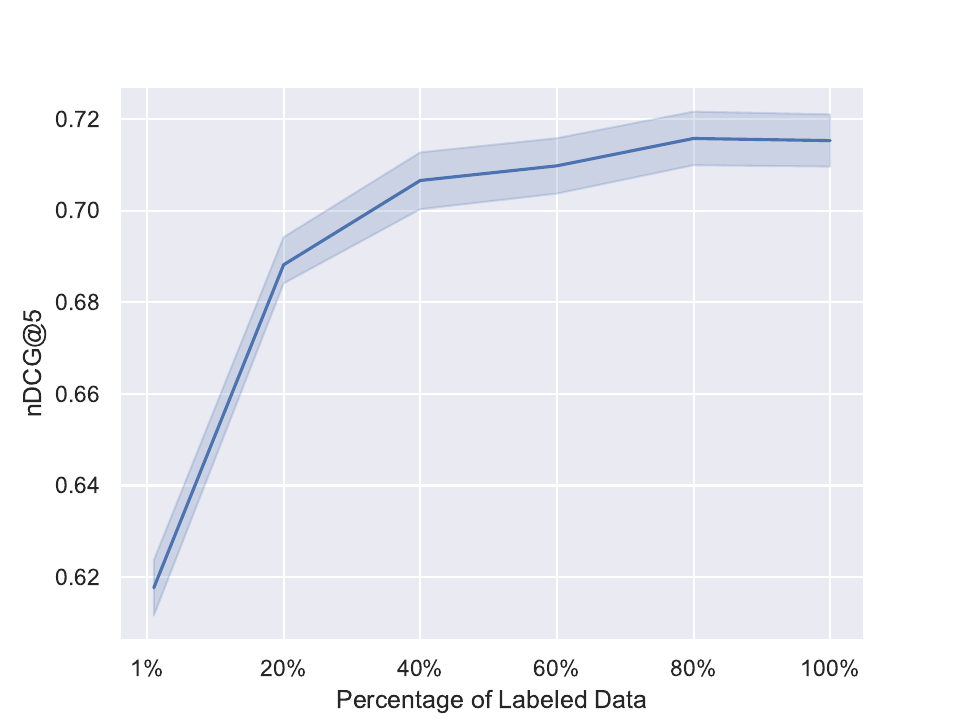}
  \vspace{-3mm}
  \caption{Comparison of production rankers trained with different percentages of labeled data on Yahoo! LETOR set.}
  \label{fig:Yahoo_label}
  \Description{Yahoo_label.}
  \vspace{-4mm}
\end{figure}

In order to use more sophisticated production rankers and align with the ranking model used in previous CLTR work \cite{ai2018dla, vardasbi2020cm-ips, chen2021iobm, Luo2024UPE}, we implemented the production rankers with deep neural networks (see Section~\ref{subsec:training} for detailed parameters). To vary the ranking performance of production rankers, we trained the DNN ranking model with 1\%, 20\%, 40\%, 60\%, and 80\% of the labeled training data, respectively. Figure~\ref{fig:Yahoo_label} shows the nDCG@5 performance of those production rankers on the Yahoo! LETOR set. Note that the production ranker trained with 100\% labeled data acts as the skyline for all the CLTR models. The performance of the production ranker needs to be worse than the skyline to leave enough room for improvement, otherwise, the CLTR models will be meaningless. Therefore, we finally chose 1\% DNN and 20\% DNN as our production rankers and used them to generate initial ranked lists for the training queries.

\subsection{User Simulation Model}
\label{section_click_simulation}
We utilized three different user simulation models, position-based model (PBM) \cite{craswell2008experimental}, dependent click model (DCM) \cite{guo2009dcm}, and comparison-based click model (CBCM) \cite{zhang2021cbcm}, to simulate clicks in three scenarios: isolated scenario, cascade scenario, and adjacency-aware scenario.

\textbf{PBM. }
In PBM, the examination on a document is only related to its position in the ranked list, so the examination and click on each document are isolated from each other. Thus, the examination hypothesis in Eq.~(\ref{eq: EH}) holds. 

Following the methodology proposed by Chapelle et al. \cite{chapelle2009expected} and Ai et al. \cite{ai2018dla}, we sampled $r_x^{q} = 1$ with:
\begin{equation}
\label{eq_relevance}
P(r_x^{q} = 1) = \epsilon + (1 - \epsilon) \frac{2^{y_x} - 1}{2^{y_{max}} - 1}
\end{equation}
where $y_x \in [0, 4]$ is the 5-level relevance label for document $x$ and
$y_{max}$ is the maximum value of $y$ (which is 4 in our case). In addition, the parameter $\epsilon$ in Eq.(\ref{eq_relevance}) is used to model click noise so that irrelevant documents ($y = 0$) have a non-zero probability of being perceived as relevant. For simplicity, we set the value of $\epsilon$ to 0.1 following Ai et al. \cite{ai2018dla} and Chen et al. \cite{chen2021iobm}.

As for sampling $o_x^{q} = 1$, we set the examination probability to:
\begin{equation}
\label{eq_PBM_exam}
    P(o_x^{q} = 1 | \pi_q) = \rho_{{\rm rank}(x|\pi_q)} = (\frac{1}{{\rm rank}(x|\pi_q)})^{\eta}
\end{equation}
where $\eta$ is a parameter that controls the severity of position bias and is set to 1. 

\textbf{DCM. } DCM is a classic cascade-based click model and also follows the examination hypothesis. Similar to PBM, we sampled $r_x^{q} = 1$ with Eq.~(\ref{eq_relevance}). In DCM, the user examines the results from top to bottom until clicking on a satisfying document. Therefore, for a document $x$, if no other document has been clicked before it in the session, then its $x$ examination probability is 1. If there is a previous click in this session, then its examination probability depends on the rank of the last clicked document:
\begin{equation}\nonumber
\label{eq_DCM_exam}
    P(o_x^{q} = 1 |  \pi_q, c_l = 1) = \lambda_l = \beta \times (\frac{1}{{\rm rank}(l|\pi_q)})^\eta
\end{equation}
where $l$ is the last clicked document before $x$ in ranked list $\pi_q$, $\beta$ and $\eta$ are parameters controlling the severity of position bias. To make DCM generate roughly the same number of clicks as PBM, we set $\beta$ and $\eta$ to 0.6 and 1, respectively.

\textbf{CBCM. } CBCM \cite{zhang2021cbcm} assumes that users would examine and compare adjacent documents to make click-through decisions, where the traditional examination hypothesis no longer holds. We set the size of the viewport to 2, and the click probability for a document in the current viewport $v$ is:
\begin{equation}\nonumber
    P(c_x^{q} = 1 |\pi_q, c_v) = 
    \frac{{\rm exp}(y_x + \rho_{{\rm rank}(x|\pi_q)} - 4\mathbb{I}(c_x^{q} = 1))}{{\rm exp}(g) + \sum\limits_{k \in v} {\rm exp}(y_k + \rho_{{\rm rank}(k|\pi_q)} - 4\mathbb{I}(c_k^{q} = 1) )  }
\end{equation}
where $\rho = (1/{\rm rank}(x|\pi_q))^\eta$ represents the position bias in the viewport, $- 4I(c_k^{q} = 1)$ is a penalty for clicked documents and $g$ controls the probability of moving downward the viewport. If the user does not click on any documents, then the viewport is moved down. After clicking a document $x$, the user has a probability of $w \times (2^{y_x} - 1) / (2^{y_{max}} - 1)$ of leaving the SERP, and otherwise return the current viewport. To generate roughly the same number of clicks as PBM and DCM, we set $g$ to 6.6 for Yahoo! and 6.55 for MSLR, set $\eta$ to 0.75, and set $w$ to 0.4.  Note that in CBCM, due to the process of comparing the relevance of adjacent documents, the click probability of a document does not follow an affine transformation of the probability that it is perceived relevant $P(r_x^{q}=1)$. According to Oosterhuis \cite{oosterhuis2022reaching}, because the click probability is not given by an affine model, the existing IPW-based CLTR models cannot theoretically achieve unbiasedness in the face of CBCM. 

\subsection{CLTR Models}
We first implemented three click baselines with different loss functions:

\textbf{click-point:} This model uses raw click data to train the ranking model with a pointwise sigmoid loss.

\textbf{click-pair:} This model uses raw click data to train the ranking model with a pairwise binary cross-entropy loss.

\textbf{click-softmax:} This model uses raw click data to train the ranking model with a listwise softmax-based cross entropy loss, the same as that used by Ai et al. \cite{ai2018dla}.

Besides, we reproduced ten prevalent CLTR models:

\textbf{IPS-PBM-EM:} Proposed by Joachims et al. \cite{joachims2017ips}, this model leverages PBM as the propensity model and computes the inverse propensity scores as Eq~(\ref{eq_ips_pbm}). Since we assumed a completely offline setting, we did not use a separate result randomization experiment to estimate the parameters of the PBM model like Joachims et al. \cite{joachims2017ips}. Instead, we utilized the expectation-maximization (EM) \cite{dempster1977EM} algorithm to estimate the parameters of PBM from the click logs.

\textbf{IPS-PBM-Reg:} This model is the same as the above IPS-PBM-EM model, except that it used Regression-EM \cite{wang2018regression-EM} to estimate the parameters of PBM from the click logs.

\textbf{IPS-DCM:} Proposed by Vardasbi et al. \cite{vardasbi2020cm-ips}, this model computes the propensities based on a DCM click model. For a document $x$, if no other document has been clicked before it
in the session, then its examination probability is 1. If there is
a previous click in this session, then its examination probability
is calculated  with:
\begin{equation}
\label{eq_DCM_propensity_computation}
    P(o_x^{q} = 1 |  \pi_q, c_l = 1) = \lambda_l,
\end{equation} 
where $l$ is the last clicked document before $x$ in ranked list $\pi_q$.  Following Vardasbi et al. \cite{vardasbi2020cm-ips}, we utilized maximum-likelihood estimation (MLE) to estimate the $\lambda$ parameters of DCM from the click logs.

\textbf{PRS-PBM-EM:} Proposed by Wang et al. \cite{wang2021prs}, the propensity ratio scoring (PRS) method integrates the inverse propensity weighting on both the clicked documents and the non-clicked ones to reweigh the pairwise losses. This model uses PBM as the propensity model as shown in Eq~(\ref{eq:prs}) and estimates its parameters with EM.

\textbf{PRS-PBM-Reg:} This model is the same as the above PRS-PBM-EM model, except that it used Regression-EM \cite{wang2018regression-EM} to estimate the parameters of PBM from the click logs.

\textbf{PRS-DCM:} This model leverages DCM to compute the propensities used in PRS and utilizes MLE to estimate the parameters of DCM.

\textbf{DLA-PBM:} Proposed by Ai et al. \cite{ai2018dla}, the DLA method jointly learns the ranking model and propensity model as stated in Section~\ref{sec:preliminaries}. This DLA model uses PBM as the propensity model.

\textbf{DLA-DCM:} This DLA model uses DCM as the propensity model and jointly learns the DCM and the ranking model.

\textbf{DLA-IOBM:} Proposed by Chen et al.\cite{chen2021iobm}, this DLA model uses the interactional observation-based model (IOBM) as the propensity model.

\textbf{UPE:} Proposed by Luo et al. \cite{Luo2024UPE}, this model augments the DLA-PBM model by leveraging the logging-policy-aware propensity (LPP) model to mitigate the confounding effect of query-document relevance in the ULTR problem.

The PRS models used the same pairwise lambda loss as Wang et al. \cite{wang2021prs} and other CLTR models used the same listwise softmax-based cross entropy loss as Ai et al. \cite{ai2018dla}. 

\subsection{Training}
\label{subsec:training}
Using the above click simulating process, we generated different numbers of sessions $\{5, 20, 100\}$ for each training query. Following Ai et al. \cite{ai2018dla}, we used deep neural networks (DNN) as the ranking model throughout the study, which have three fully connected layers with sizes $\{512, 256, 128\}$, an $elu$ activation function, and layer normalization. When training the CLTR models, only the top 10 documents were considered to be displayed. This setting is the same as Ai et al.\cite{ai2018dla} to simulate the top-K truncation in real-world display scenarios \cite{Baidu} and introduces item selection bias \cite{oosterhuis2020policy-aware, ovaisi2020selectionbias}. Since we use deterministic production rankers, the methods designed to deal with selection bias by Oosterhuis et al. \cite{oosterhuis2020policy-aware} and Ovaisi et al. \cite{ovaisi2020selectionbias} cannot be reproduced. We set the batch size to 256 and set the learning rate to 0.01 for Yahoo! LETOR and 0.1 for MSLR-WEB10K. We trained each CLTR model for 10K steps and adopted the hyperparameters with the best nDCG@5 results on the validation set. Note that under our simulation settings, the CLTR methods are no longer strictly unbiased because there exists a selection bias and the examination propensities are estimated offline.

\subsection{Evaluation Using a New Metric}
We evaluated the performance of CLTR models on the test set of Yahoo! LETOR and MSLR-WEB10K with human relevance annotations. We used normalized Discounted Cumulative Gain (nDCG) \cite{jarvelin2002cumulated} to measure the performance of the ranking models. Since a CLTR model aims to learn a new ranking model from historical click logs with performance improvement over the existing production ranker, we compute the normalized nDCG@5 increase rates of CLTR models against the production ranker to measure and compare their effectiveness under different simulation settings:
\begin{align}
\begin{split}
    &\rm Inc({\it{M}}) = \frac{nDCG@5({\it{M}}) - nDCG@5(PR)}{nDCG@5(PR)} \quad ,\\
    &\rm nInc({\it{M}}) = \frac{Inc({\it{M}})}{Inc(Skyline)} \quad,
\end{split}
\end{align}
where $M$ represents a CLTR model, ``PR'' represents the production ranker, and the ``Skyline'' refers to the DNN model trained with 100\% labeled training data. 

\subsection{Code and Data}
We release the code of the complete pipeline, including generating ranked lists, simulating clicks, and training the CLTR models. To facilitate the reproducibility of the reported results, our study only made use of publicly available datasets. All the related resources with instructions can be accessed at the github repository \href{https://github.com/Diligentspring/Investigating-the-Robustness-of-Counterfactual-Learning-to-Rank-Models}{github.com/Diligentspring/Investigating-the-Robustness-of-Counterfactual-Learning-to-Rank-Models}.\footnote{\href{https://github.com/Diligentspring/Investigating-the-Robustness-of-Counterfactual-Learning-to-Rank-Models}{https://github.com/Diligentspring/Investigating-the-Robustness-of-Counterfactual-Learning-to-Rank-Models}}

\section{Results and Analysis}
\label{sec:results}
In this section, we report and discuss the results of our extended simulation-based experiment. In particular, we focus on the following research questions:

\begin{figure*}[ht]
  \centering
  \includegraphics[width=0.87\linewidth,trim=140 10 100 50,clip]{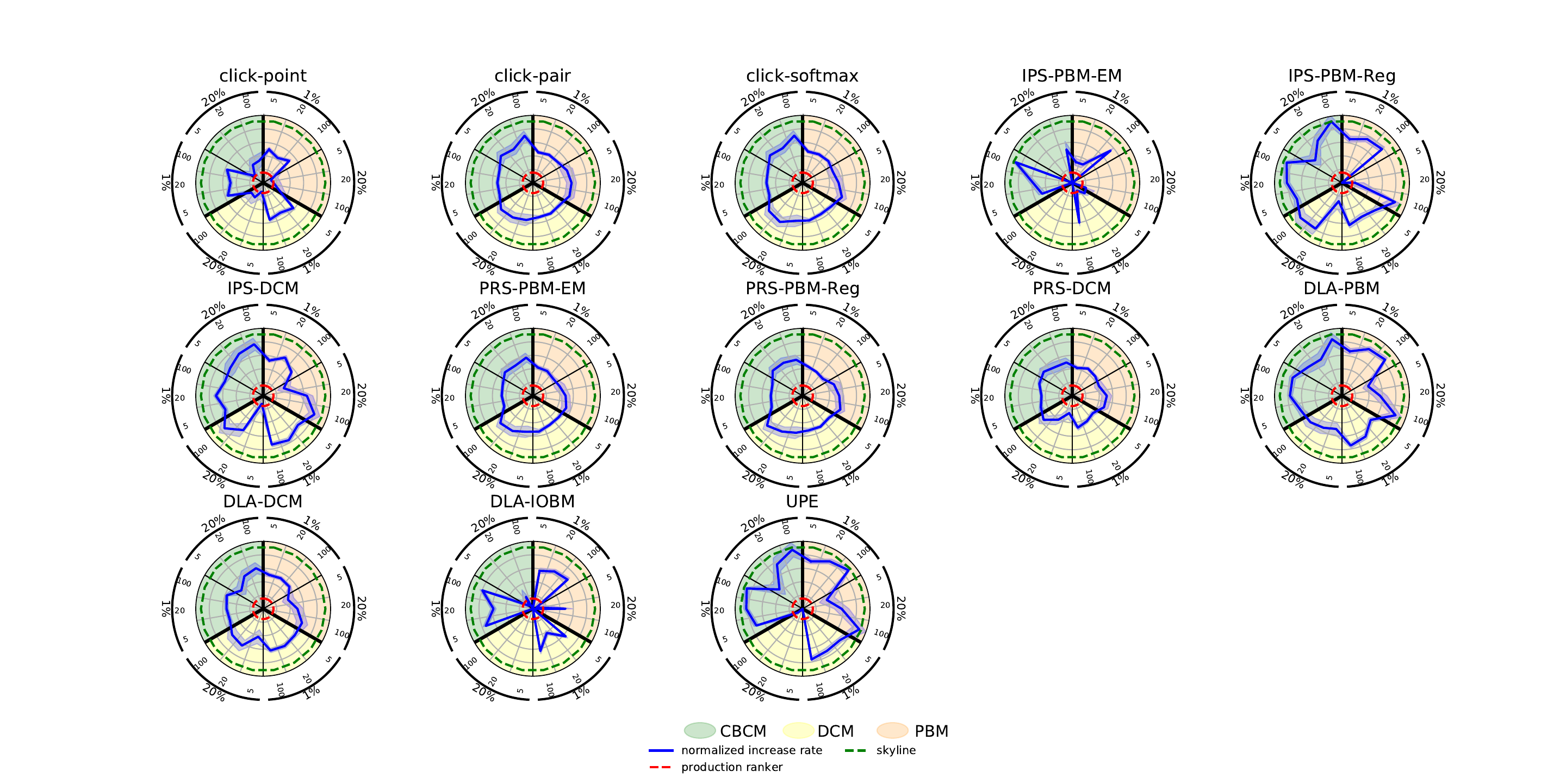}
  \vspace{-4mm}
  \caption{Spider charts showing the level of robustness in a wide range of simulated environments on Yahoo! LETOR. The increase rates represented
on each graph are normalized increase rates where the best model is close to the border and the worst model is close to the center. The background colors represent the type of user simulation model, the outer circle of labels indicates the percentage of data used to train the production ranker, and the inner circle indicates the number of synthetic sessions used to train the CLTR models. 95\% confidence bounds appear in shaded areas.}
  \label{fig:Yahoo_robustness}
  \vspace{-4mm}
  \Description{This figure shows the robustness.}
\end{figure*}

\begin{figure*}[ht]
  \centering
  \includegraphics[width=0.87\linewidth,trim=140 10 100 50,clip]{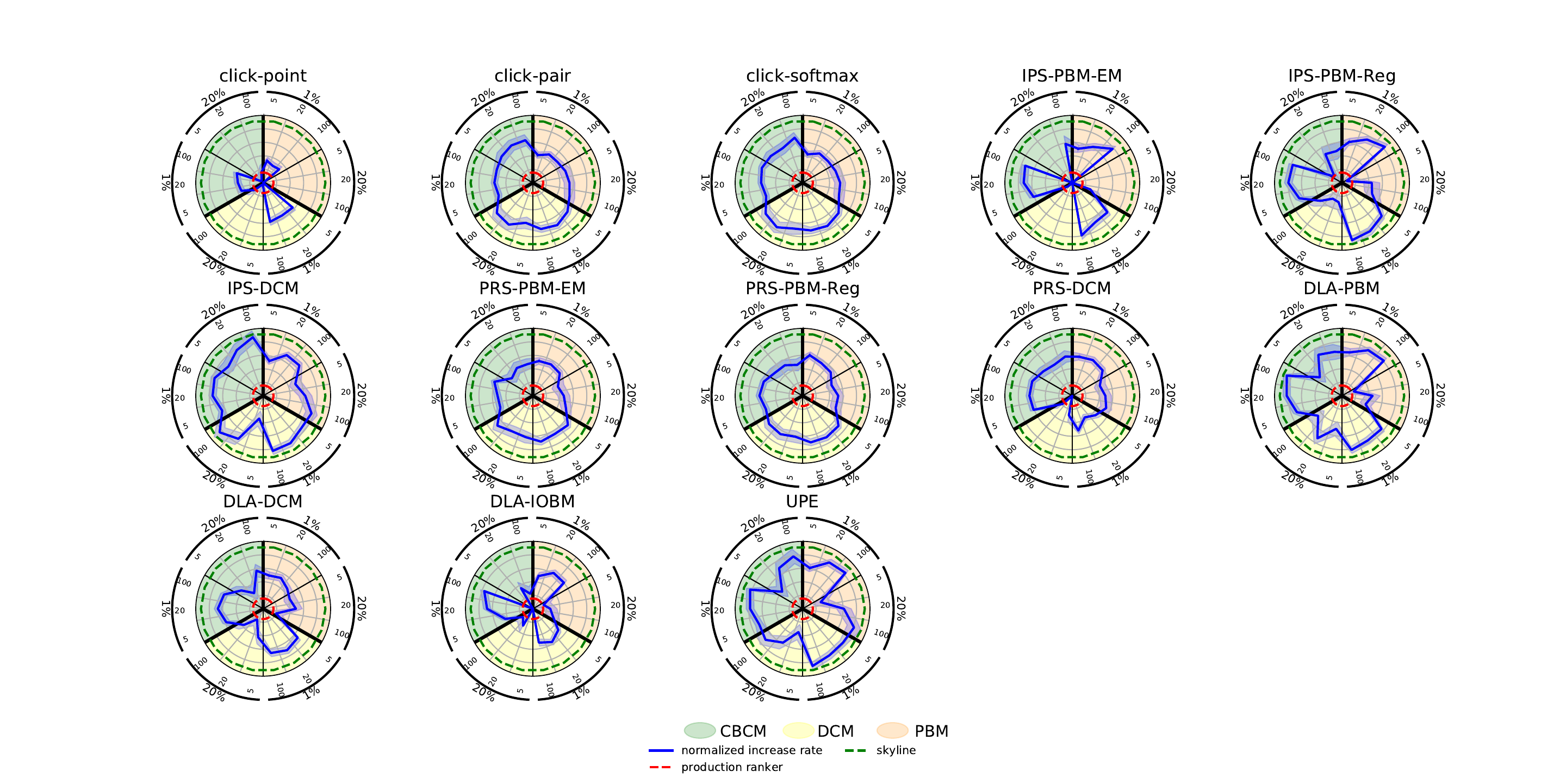}
  \vspace{-4mm}
  \caption{Spider charts showing the level of robustness in a wide range of simulated environments on MSLR-WEB10K. The increase rates represented
  on each graph are normalized increase rates where the best model is close to the border and the worst model is close to the center. The background colors represent the type of user simulation model, the outer circle of labels indicates the percentage of data used to train the production ranker, and the inner circle indicates the number of synthetic sessions used to train the CLTR models. 95\% confidence bounds appear in shaded areas.}
  \label{fig:MSLR_robustness}
  \vspace{-4mm}
  \Description{This figure shows the robustness.}
\end{figure*}

\subsection{RQ1: How Robust Are the CLTR Models Under Various Simulation Settings?}
\label{sec:RQ1}
Figure~\ref{fig:Yahoo_robustness} and Figure~\ref{fig:MSLR_robustness} show the normalized nDCG@5 increase rate (nInc) performance of the CLTR models under various simulation settings on Yahoo! LETOR and MSLR, respectively.\footnote{In order to compare the robustness of each CLTR model more clearly in the spider charts, we set the minimum value of nInc to -0.2.} The closer the blue line in the spider chart is to the green dotted line, the better the performance of the model is. If the blue line is wrapped by the red dotted line, it means that the model cannot surpass the production ranker. For the CLTR methods, we can see that the patterns formed by the blue lines of IPS-DCM, DLA-PBM, and UPE are rounder and cover larger areas than others. Therefore, we have \textbf{Finding~(1): the IPS-DCM, DLA-PBM, and UPE models show better robustness under various simulation settings than other CLTR models and the click baselines.} We argue that DLA-PBM and UPE benefit from the simplicity of PBM assumptions and the ability of the dual learning algorithm (DLA) to adaptively adjust the propensity estimation. As for IPS-DCM, the reason for its robustness may be the unique way it calculates propensities. When there is only one click in a session, IPS-DCM assigns a propensity of one for the clicked document, which is equivalent to the click-softmax baseline and would be less affected by a possible noise in a low position.

For the three naive baselines, the patterns formed by the blue lines of click-pair and click-softmax are closer to a circle and cover a larger area than those of click-point and IPS-PBM-EM. Thus, we have \textbf{Finding~(2): the click-pair and click-softmax baselines achieve relatively good overall performance and even beat some CLTR models.} Interestingly, the phenomenon that some CLTR models cannot outperform the click baselines also appeared in the experiments conducted on real-world click logs by Zou et al. \cite{Baidu} and Hager et al. \cite{hager2024unbiased}. Moreover, the performance of click-pair and click-softmax is significantly better than click-point, which shows that pairwise and listwise loss functions have advantages over the pointwise loss function and future CLTR work should use the former two to implement the naive click baselines. 

In terms of propensity estimation methods, IPS-PBM-Reg and PRS-PBM-Reg outperform their counterparts IPS-PBM-EM and PRS-EM, respectively. Thus, we have \textbf{Finding~(3): the CLTR models using Regression-EM to estimate the parameters of PBM perform better than those using EM.} This indicates the superiority of the Regression-EM algorithm over the traditional EM algorithm in estimating the position bias.

\begin{table}[t]
  \caption{nInc performance of CLTR models on Yahoo! LETOR with 20\% DNN as the production ranker and 5 synthetic sessions for each training query. The best-performing model is in bold. The best-performing click baseline is underlined. $\uparrow/\downarrow$ indicates a result is significantly better or worse (Tukey's HSD test with p-value < 0.05) than the best-performing click baseline.}
  \label{tab:drawback_Yahoo}
  \vspace{-4mm}
  \begin{tabular}{cccc} 
\toprule
CLTR model& PBM & DCM & CBCM \\ \midrule
click-point& -0.0414$\downarrow$ & -0.0277$\downarrow$ & 0.0075$\downarrow$ \\ 
click-pair&	\textbf{\underline{0.4991}} & 0.5290 & 0.6058 \\ 
click-softmax& 0.4213 &	\textbf{\underline{0.5450}} & \underline{0.6215} \\ 
IPS-PBM-EM\cite{joachims2017ips, dempster1977EM}& -3.2192$\downarrow$ &	-4.5531$\downarrow$ & -3.3614$\downarrow$\\
IPS-PBM-Reg\cite{joachims2017ips, wang2018regression-EM}& -0.3398$\downarrow$ &	0.1604$\downarrow$ & 0.4722\\
IPS-DCM\cite{vardasbi2020cm-ips}& 0.2238&	-0.0441$\downarrow$ & 0.6307\\ 
PRS-PBM-EM\cite{wang2021prs, dempster1977EM}& 0.3471 &	0.5048 & 0.5040 \\ 
PRS-PBM-Reg\cite{wang2021prs, wang2018regression-EM}& 0.4430 & 0.5240 & 0.5518\\ 
PRS-DCM& 0.3480 & 0.1390$\downarrow$ & 0.5188 \\ 
DLA-PBM\cite{ai2018dla}& 0.3356 &	0.4496 & \textbf{0.6567} \\ 
DLA-DCM& 0.3095 & 0.3465 & 0.3495 \\ 
DLA-IOBM\cite{chen2021iobm}& -0.1254$\downarrow$ &  -0.8498$\downarrow$ & -0.1057$\downarrow$ \\
UPE\cite{Luo2024UPE}& 0.2966 & -0.3575$\downarrow$ & 0.3807\\
\bottomrule
\end{tabular}
\vspace{-4mm}
\end{table}

In addition, we observe that the CLTR models often fail to outperform the naive click-pair and click-softmax baselines when the production ranker is 20\% DNN and the number of synthetic click logs is limited. \footnote{Please note this phenomenon is different from the ``unsafety'' problem discussed in \cite{Gupta2023SafeCLTR, Gupta2024PRPO} that CLTR models can perform worse than the production ranker when the available click data are sparse.} To demonstrate this phenomenon more clearly, we show the nInc performance of the CLTR models under this setting in Table~\ref{tab:drawback_Yahoo}. We have \textbf{Finding~(4): the CLTR models often fail to outperform the best-performing naive baseline when the production ranker has relatively high ranking performance (20\% DNN) and the number of training sessions is relatively small (5 sessions for each query).} This finding warns us to be careful when using CLTR models in real-world scenarios where the production ranker has a good ranking performance and click data are sparse. In addition, it suggests an urgent need to develop new CLTR algorithms that work in these settings.

\subsection{RQ2: How Does the Production Ranker Affect the CLTR Models?}

\begin{table}[t]
  \caption{Analysis of variance (ANOVA) on Yahoo! LETOR. The columns from left to right show the source of variance, degree of freedom, sum of squared deviations, mean square, F statistics, p-value, and effect size. The analyzed factors include the performance of production ranker (1\% DNN and 20\% DNN), number of synthetic sessions, user simulation model, CLTR model, and residual.}
  \label{tab:anova_Yahoo}
  \vspace{-3mm}
  \resizebox{0.45\textwidth}{!}{
  \begin{tabular}{lcccccc} 
\toprule
Source & SS & DF & MS & F & $p$-value & $\hat{\omega}^{2}$ \\ \midrule
Performance of PR& 7.6361& 1&	7.6361&	25.3424&	1.01e-6&	0.0942 \\ 
User simulation model& 0.7826 &	2& 1.5651&	2.5971&	7.68e-2&	0.0135\\ 
Number of sessions&	 1.6399	& 2& 3.2799&	5.4426&  4.94e-3&	0.0366\\ 
CLTR model& 3.4640&	12& 41.5685&	11.4963&	9.28e-18&	0.3499\\ 
Residual& 0.3013&	216 & 65.0847& &	& \\ \bottomrule
\end{tabular}
}
\vspace{-3mm}
\end{table}

Before answering the RQ2, we first conducted analysis of variance (ANOVA) on how the factors involved in our simulation-based experiment affect the nInc performance of the CLTR models on Yahoo! LETOR. As shown in Table~\ref{tab:anova_Yahoo}, all the effects are significant ($p<0.05$), where ``CLTR model'' has a large effect size ($\hat{\omega}^{2} \geq 0.14$), ``Performance of PR'' has a medium size effect ($0.06 \leq \hat{\omega}^{2} < 0.14$), and ``User simulation model'' and ``Number of sessions'' each has a small effect size ($0.01 \leq \hat{\omega}^{2} < 0.06$).

\begin{table}[t]
  \caption{nInc performance of CLTR models on Yahoo! LETOR with different production rankers. The best-performing model is in bold. The best-performing click baseline is underlined. $\uparrow/\downarrow$ indicates a result is significantly better or worse (Tukey's HSD test with p-value < 0.05) than the best-performing click baseline.}
  \label{tab:production_ranker_Yahoo}
  \vspace{-3mm}
  \begin{tabular}{ccc} 
\toprule
CLTR model& 1\% DNN & 20\% DNN \\ 
\midrule
\multicolumn{3}{c}{using PBM} \\
\hline
click-point&  0.4214 & 0.0170$\downarrow$\\ 
click-pair&	0.4144 & \underline{0.5318}\\ 
click-softmax& \underline{0.4315} & 0.5080\\ 
IPS-PBM-EM\cite{joachims2017ips, dempster1977EM}& 	0.3933 & -2.0825$\downarrow$\\
IPS-PBM-Reg\cite{joachims2017ips, wang2018regression-EM}& 	0.7454$\uparrow$ & 0.2009$\downarrow$\\
PRS-PBM-EM\cite{wang2021prs, dempster1977EM}& 0.3383$\downarrow$ & 0.4210\\ 
PRS-PBM-Reg\cite{wang2021prs, wang2018regression-EM}& 0.3393$\downarrow$ & 0.5086\\ 
DLA-PBM\cite{ai2018dla}& 	0.7982$\uparrow$ & 0.5900\\ 
UPE\cite{Luo2024UPE}& \textbf{0.8461}$\uparrow$ &  \textbf{0.6041}\\
\hline
\multicolumn{3}{c}{using DCM} \\
\hline
click-point&  0.5158 & 0.0640$\downarrow$ \\ 
click-pair&	0.4756 & 0.5670\\ 
click-softmax& 	\underline{0.5167} & \textbf{\underline{0.6206}}\\ 
IPS-DCM\cite{vardasbi2020cm-ips}& \textbf{0.7400}$\uparrow$ & 0.4315$\downarrow$\\ 
PRS-DCM& 0.3698$\downarrow$ & 0.3310$\downarrow$\\ 
DLA-DCM& 0.6135$\uparrow$	& 0.5160 \\ 
\bottomrule
\end{tabular}
\vspace{-4mm}
\end{table}

Next, we compared the nInc performance of CLTR models under different production ranker settings. We controlled the user simulation model to be consistent with the propensity model used by the CLTR models to avoid the interference of ``click model mismatch".\footnote{Click model mismatch refers to the mismatch between the propensity model used by the CLTR method and the user simulation model used for simulating clicks.} Besides, we calculated the average performance of the CLTR models over different numbers of synthetic sessions to comprehensively consider the influence of variance. The results on Yahoo! LETOR are shown in Table~\ref{tab:production_ranker_Yahoo}. 

When the production ranker changes from 1\% DNN to 20\% DNN, we see that the performance of click-pair and click-softmax increases, while the performance of the IPS and DLA series of CLTR models decreases. Thus, we have \textbf{Finding~(5): when the performance of the production ranker is higher, the click-pair and click-softmax baselines perform better, while the IPS and DLA series of CLTR models tend to perform worse.} When the production ranker becomes stronger, relevant documents are usually ranked higher and thus more likely to receive more observations and clicks than irrelevant documents, so the click-pair and click-softmax have better performance. As for the performance degradation of the CLTR models, it may be due to the confounding effect of the relevance in estimating the propensities as pointed out by Luo et al. \cite{Luo2024UPE}. Besides, it may be because the IPW-based methods would be more sensitive to the noises at the lower ranks when facing strong production rankers. In addition, we observe that UPE shows better performance than DLA-PBM when using the DNN production rankers, which verifies the superiority of UPE over DLA-PBM when facing strong production rankers.

\subsection{RQ3: How Does the User Simulation Model Affect the CLTR Models?}
To compare the performance of CLTR models under different user simulation model settings, we controlled the production ranker to be the weaker 1\% DNN to avoid the interference of a ``strong production ranker" and calculated the average nInc values over different numbers of synthetic sessions. The results on Yahoo! LETOR are shown in Table~\ref{tab:user simulation model}.

\begin{table}[t]
  \caption{nInc performance of CLTR models on Yahoo! LETOR with 1\% DNN as the production ranker and different user simulation models. The best-performing model is in bold. The best-performing click baseline is underlined. $\uparrow/\downarrow$ indicates a result is significantly better or worse (Tukey's HSD test with p-value < 0.05) than the best-performing click baseline.}
  \label{tab:user simulation model}
  \vspace{-3mm}
  \begin{tabular}{cccc} 
\toprule
CLTR model& PBM & DCM & CBCM \\ \midrule
click-point& 0.4214 & 0.5158 & \underline{0.4981} \\ 
click-pair&	0.4144 &	0.4756 & 0.4837 \\ 
click-softmax& \underline{0.4315} &	\underline{0.5167} & 0.4968 \\ 
IPS-PBM-EM\cite{joachims2017ips, dempster1977EM}& 0.3933 &	0.2229$\downarrow$ & 0.6462$\uparrow$\\
IPS-PBM-Reg\cite{joachims2017ips, wang2018regression-EM}& 0.7454$\uparrow$ &	0.6039$\uparrow$ & 0.8397$\uparrow$\\
IPS-DCM\cite{vardasbi2020cm-ips}& 0.5512$\uparrow$&	0.7400$\uparrow$ & 0.6248$\uparrow$\\ 
PRS-PBM-EM\cite{wang2021prs, dempster1977EM}& 0.3383$\downarrow$ &	0.4645 & 0.4021$\downarrow$ \\ 
PRS-PBM-Reg\cite{wang2021prs, wang2018regression-EM}& 0.3393$\downarrow$ &	0.4763 & 0.4217$\downarrow$\\ 
PRS-DCM& 0.3768 & 0.3698$\downarrow$ & 0.4377$\downarrow$ \\ 
DLA-PBM\cite{ai2018dla}& 0.7982$\uparrow$ &	0.6681$\uparrow$ & 0.7521$\uparrow$ \\ 
DLA-DCM& 0.4676 & 0.6135$\uparrow$ & 0.5118 \\ 
DLA-IOBM\cite{chen2021iobm}& 0.6156$\uparrow$&  0.5313 & 0.7204$\uparrow$ \\
UPE\cite{Luo2024UPE}& \textbf{0.8461}$\uparrow$ & \textbf{0.7574}$\uparrow$ & \textbf{0.8594}$\uparrow$\\
\bottomrule
\end{tabular}
\vspace{-1mm}
\end{table}

All the IPS and DLA series of CLTR models except IPS-PBM-EM robustly outperform the click baselines on both the PBM and DCM simulation models no matter whether there exists a click model mismatch. Even in the face of the CBCM simulation model, which is the most unsound case in theory according to Oosterhuis \cite{oosterhuis2022reaching}, all the IPS and DLA models outperform the click baselines and some (IPS-PBM, DLA-IOBM, and UPE) even achieve their best performance. Therefore, we have \textbf{Finding~(6): although theoretically unsound, the IPS and DLA series of CLTR models show certain robustness to click model mismatch.} This finding is consistent with the small effect size of ``User simulation model'' shown in Table~\ref{tab:anova_Yahoo}.

\subsection{RQ4: How Does the Size of Synthetic Click Logs Affect the CLTR Models?}

\begin{table}[t]
  \caption{nInc performance of CLTR models on Yahoo! LETOR with different sizes of synthetic click logs. The best-performing model is in bold. The best-performing click baseline is underlined. $\uparrow/\downarrow$ indicates a result is significantly better or worse (Tukey's HSD test with p-value < 0.05) than the best-performing click baseline.}
  \label{tab:number_sessions}
  \vspace{-3mm}
  \begin{tabular}{cccc} 
\toprule
\multirow{2}{*}{CLTR model}& \multicolumn{3}{c}{Number of sessions per query} \\
\cline{2-4}
 & 5 & 20 & 100 \\
\midrule
\multicolumn{4}{c}{using PBM} \\
\hline
click-point&  \underline{0.4527} & 0.3544 & \underline{0.4572} \\ 
click-pair&	0.3970 & 0.4231 & 0.4233\\ 
click-softmax& 0.4119 & \underline{0.4326} & 0.4499\\ 
IPS-PBM-EM\cite{joachims2017ips, dempster1977EM}& 	0.2052$\downarrow$ & 0.2070$\downarrow$ &0.7676$\uparrow$\\
IPS-PBM-Reg\cite{joachims2017ips, wang2018regression-EM}& 	0.6535$\uparrow$ & 0.7729$\uparrow$ & 0.8097$\uparrow$\\
PRS-PBM-EM\cite{wang2021prs, dempster1977EM}& 0.3540 & 0.3504 & 0.3104$\downarrow$\\ 
PRS-PBM-Reg\cite{wang2021prs, wang2018regression-EM}& 0.3754 & 0.3327 & 0.3097$\downarrow$\\ 
DLA-PBM\cite{ai2018dla}& 	0.6706$\uparrow$ & 0.8396$\uparrow$ &0.8845$\uparrow$\\ 
UPE\cite{Luo2024UPE}& \textbf{0.7278}$\uparrow$ &  \textbf{0.8548}$\uparrow$ & \textbf{0.9557}$\uparrow$\\
\hline
\multicolumn{4}{c}{using DCM} \\
\hline
click-point&  \underline{0.5586} & 0.4653 & 0.5234\\ 
click-pair&	0.4704$\downarrow$ & 0.4838 & 0.4725\\ 
click-softmax& 	0.5136 & \underline{0.4990} & \underline{0.5376}\\ 
IPS-DCM\cite{vardasbi2020cm-ips}& \textbf{0.6758}$\uparrow$ & \textbf{0.7885}$\uparrow$ & \textbf{0.7556}$\uparrow$\\ 
PRS-DCM& 0.3191$\downarrow$ & 0.3663$\downarrow$ & 0.4240$\downarrow$\\ 
DLA-DCM& 0.5950	& 0.6283$\uparrow$ & 0.6172 \\ 
\bottomrule
\end{tabular}
\vspace{-4mm}
\end{table}

Finally, we investigate how the CLTR models perform in the face of different sizes of synthetic click logs. We controlled the production ranker to be the weaker 1\% DNN to avoid the interference of a ``strong production ranker". Besides, we controlled the user simulation model to be consistent with the propensity model used by the CLTR models to avoid the interference of ``click model mismatch". The results on Yahoo! LETOR are shown in Table~\ref{tab:number_sessions}. We have \textbf{Finding~(7): the performance of the IPS and DLA series of CLTR models tends to decrease as the number of synthetic click logs decreases.} This finding is as expected and consistent with the results of Joachims et al. \cite{joachims2017ips} because the IPS-based and DLA-based CLTR models suffer from higher variance when the click data are sparser.

\section{Related Work}
\label{sec:relatedwork}

In this section, we review some online learning to rank models and other user behavior models that are not covered by our study.

\textbf{Online Learning to rank models. }
Different from offline CLTR models, online learning to rank (OLTR) models make online interventions of ranked lists and learn from real-time user interactions. That is, OLTR models decide what ranked lists to display to users and update the parameters using the interactions instantly, which is out of the scope of our study. 

The dual bandit gradient descent (DBGD) model proposed by Yue and Joachims \cite{yue2009DBGD} is one of the most classical OLTR models, which iteratively creates random parameter perturbations and uses an interleaved test \cite{chapelle2012large} to find the optimal perturbed parameters to update the ranking model. Later researches extended DBGD by developing more efficient sampling strategies \cite{schuth2016MGD, wang2018efficient, zhao2016constructing} and variance reduction techniques \cite{wang2019variance}. For example, Schuth et al. \cite{schuth2016MGD} proposed to simultaneously compare multiple perturbed parameters using multileaved comparison methods \cite{schuth2014multileaved}. Oosterhuis and de Rijke \cite{oosterhuis2018PDGD} proposed pairwise differentiable
gradient descent (PDGD) that constructs a weighted differentiable pairwise loss based on inferring preferences between document pairs from user clicks to avoid the interleaving or multileaving experiments.

\textbf{User behavior models. }
Our study only used three representative click models, while there are many other click models with different user behavior assumptions. For example, the user browsing model (UBM) proposed by Dupret and Piwowarski \cite{dupret2008ubm} is a combination of the PBM and CM click model, which assumes that the examination probability of a document is related to both its position and the position of the last clicked document. The dynamic Bayesian network model (DBN) \cite{chapelle2009DBN} further extends the cascade model, and introduces a set of document-dependent parameters to better model the probability that a user is satisfied after clicking on a document. The mobile click model (MCM) \cite{mao2018MCM} incorporates the related click necessity bias and examination satisfaction bias, and can be regarded as a unified generalization of DBN and UBM. 

The above click models are based on a probabilistic graphical model (PGM), while some other click models utilize more expressive neural networks to model user behaviors \cite{borisov2016neural, borisov2018click, chen2020context}. For example, Borisov et al. \cite{borisov2016neural} used a vector state to represent the user’s information need and the information available to the user during search, and utilized a recurrent neural network to describe the user's cascading browsing behavior. Borisov et al. \cite{borisov2018click} further proposed a click sequence model (CSM) that aims to predict the order in which a user interacts with the documents. Based on an encoder-decoder architecture, CSM first encodes contextual embeddings of the documents, and then decodes the sequence of positions of the clicked documents. In addition, Chen et al. \cite{chen2020context} proposed a context-aware click model (CACM) based on deep neural networks that leverages the behavior signals in previous iterations to better model user behaviors.

\section{Conclusion and Discussion}
\label{sec:conclusion}
In this work, we reproduced ten classic counterfactual learning to rank models and extended the mainstream simulation-based experiments with production rankers of different ranking performance, more types of user simulation models, and different sizes of click logs to further investigate the robustness of CLTR models in complex and diverse situations. 

\subsection{Main Findings}
Importantly, we have seven interesting findings from our extended simulation-based experiments. Finding~(1): the IPS-DCM, DLA-PBM, and UPE models show better robustness under various simulation settings than other CLTR models and the click baselines. Finding~(2): the click-pair and click-softmax baselines achieve relatively good overall performance and even beat some CLTR models. This finding indicates that future CLTR work should use the pairwise or listwise loss function to implement the naive click baseline. Finding~(3): the CLTR models using Regression-EM to estimate the parameters of PBM perform better than those using EM. Finding~(4): the CLTR models often fail to outperform the best-performing naive baseline when the production ranker has relatively high ranking performance (20\% DNN) and the number of training sessions is relatively small (5 sessions for each query). This finding suggests an urgent need for developing new CLTR algorithms tailored to these conditions. Finding~(5): when the performance of the production ranker is higher, the click-pair and click-softmax baselines perform better, while the IPS and DLA series of CLTR models tend to perform worse. Finding~(6): although theoretically unsound, the IPS and DLA series of CLTR models show certain robustness to click model mismatch. Finding~(7): the performance of the IPS and DLA series of CLTR models tends to decrease as the number of synthetic click logs decreases.

\subsection{Limitations and Future Work}
In this paper, we only made use of PBM, DCM, and CBCM as the user simulation model to simulate three different kinds of scenarios. However, real user behavior patterns in the real world can be more diverse and complicated. In the future, our simulation-based experiment can be extended to include more user simulation models such as UBM, DBN, MCM, and those neural click models, and even a mixture of these models. In addition, due to the lack of query frequency information, we generated the same number of sessions for each training query in our simulation-based experiments. In the future, we would like to investigate how the CLTR models perform when the training queries have different frequencies.

\begin{acks}
This research was supported by the Natural Science Foundation of China (61902209, 62377044), Intelligent Social Governance Platform, Major Innovation \& Planning Interdisciplinary Platform for the ``Double-First Class" Initiative, Renmin University of China, the Fundamental Research Funds for the Central Universities, and the Research Funds of Renmin University of China (22XNKJ15). Any opinions, findings, conclusions, or recommendations expressed in this material are those of the authors and do not necessarily reflect those of the sponsors.
\end{acks}

\bibliographystyle{ACM-Reference-Format}
\balance
\bibliography{sample-base}


\begin{thebibliography}{48}


\ifx \showCODEN    \undefined \def \showCODEN     #1{\unskip}     \fi
\ifx \showISBNx    \undefined \def \showISBNx     #1{\unskip}     \fi
\ifx \showISBNxiii \undefined \def \showISBNxiii  #1{\unskip}     \fi
\ifx \showISSN     \undefined \def \showISSN      #1{\unskip}     \fi
\ifx \showLCCN     \undefined \def \showLCCN      #1{\unskip}     \fi
\ifx \shownote     \undefined \def \shownote      #1{#1}          \fi
\ifx \showarticletitle \undefined \def \showarticletitle #1{#1}   \fi
\ifx \showURL      \undefined \def \showURL       {\relax}        \fi
\providecommand\bibfield[2]{#2}
\providecommand\bibinfo[2]{#2}
\providecommand\natexlab[1]{#1}
\providecommand\showeprint[2][]{arXiv:#2}

\bibitem[Agarwal et~al\mbox{.}(2019)]%
        {agarwal2019general}
\bibfield{author}{\bibinfo{person}{Aman Agarwal}, \bibinfo{person}{Kenta Takatsu}, \bibinfo{person}{Ivan Zaitsev}, {and} \bibinfo{person}{Thorsten Joachims}.} \bibinfo{year}{2019}\natexlab{}.
\newblock \showarticletitle{A general framework for counterfactual learning-to-rank}. In \bibinfo{booktitle}{\emph{Proceedings of the 42nd International ACM SIGIR Conference on Research and Development in Information Retrieval}}. \bibinfo{pages}{5--14}.
\newblock


\bibitem[Ai et~al\mbox{.}(2018)]%
        {ai2018dla}
\bibfield{author}{\bibinfo{person}{Qingyao Ai}, \bibinfo{person}{Keping Bi}, \bibinfo{person}{Cheng Luo}, \bibinfo{person}{Jiafeng Guo}, {and} \bibinfo{person}{W~Bruce Croft}.} \bibinfo{year}{2018}\natexlab{}.
\newblock \showarticletitle{Unbiased learning to rank with unbiased propensity estimation}. In \bibinfo{booktitle}{\emph{The 41st International ACM SIGIR Conference on Research \& Development in Information Retrieval}}. \bibinfo{pages}{385--394}.
\newblock


\bibitem[Borisov et~al\mbox{.}(2016)]%
        {borisov2016neural}
\bibfield{author}{\bibinfo{person}{Alexey Borisov}, \bibinfo{person}{Ilya Markov}, \bibinfo{person}{Maarten De~Rijke}, {and} \bibinfo{person}{Pavel Serdyukov}.} \bibinfo{year}{2016}\natexlab{}.
\newblock \showarticletitle{A neural click model for web search}. In \bibinfo{booktitle}{\emph{Proceedings of the 25th International Conference on World Wide Web}}. \bibinfo{pages}{531--541}.
\newblock


\bibitem[Borisov et~al\mbox{.}(2018)]%
        {borisov2018click}
\bibfield{author}{\bibinfo{person}{Alexey Borisov}, \bibinfo{person}{Martijn Wardenaar}, \bibinfo{person}{Ilya Markov}, {and} \bibinfo{person}{Maarten De~Rijke}.} \bibinfo{year}{2018}\natexlab{}.
\newblock \showarticletitle{A click sequence model for web search}. In \bibinfo{booktitle}{\emph{The 41st International ACM SIGIR Conference on Research \& Development in Information Retrieval}}. \bibinfo{pages}{45--54}.
\newblock


\bibitem[Chapelle and Chang(2011)]%
        {Yahoo!}
\bibfield{author}{\bibinfo{person}{Olivier Chapelle} {and} \bibinfo{person}{Yi Chang}.} \bibinfo{year}{2011}\natexlab{}.
\newblock \showarticletitle{Yahoo! Learning to Rank Challenge Overview}. In \bibinfo{booktitle}{\emph{Proceedings of the Yahoo! Learning to Rank Challenge}}. \bibinfo{pages}{1--24}.
\newblock


\bibitem[Chapelle et~al\mbox{.}(2012)]%
        {chapelle2012large}
\bibfield{author}{\bibinfo{person}{Olivier Chapelle}, \bibinfo{person}{Thorsten Joachims}, \bibinfo{person}{Filip Radlinski}, {and} \bibinfo{person}{Yisong Yue}.} \bibinfo{year}{2012}\natexlab{}.
\newblock \showarticletitle{Large-scale validation and analysis of interleaved search evaluation}.
\newblock \bibinfo{journal}{\emph{ACM Transactions on Information Systems (TOIS)}} \bibinfo{volume}{30}, \bibinfo{number}{1} (\bibinfo{year}{2012}), \bibinfo{pages}{1--41}.
\newblock


\bibitem[Chapelle et~al\mbox{.}(2009)]%
        {chapelle2009expected}
\bibfield{author}{\bibinfo{person}{Olivier Chapelle}, \bibinfo{person}{Donald Metlzer}, \bibinfo{person}{Ya Zhang}, {and} \bibinfo{person}{Pierre Grinspan}.} \bibinfo{year}{2009}\natexlab{}.
\newblock \showarticletitle{Expected reciprocal rank for graded relevance}. In \bibinfo{booktitle}{\emph{Proceedings of the 18th ACM conference on Information and knowledge management}}. \bibinfo{pages}{621--630}.
\newblock


\bibitem[Chapelle and Zhang(2009)]%
        {chapelle2009DBN}
\bibfield{author}{\bibinfo{person}{Olivier Chapelle} {and} \bibinfo{person}{Ya Zhang}.} \bibinfo{year}{2009}\natexlab{}.
\newblock \showarticletitle{A dynamic bayesian network click model for web search ranking}. In \bibinfo{booktitle}{\emph{Proceedings of the 18th international conference on World wide web}}. \bibinfo{pages}{1--10}.
\newblock


\bibitem[Chen et~al\mbox{.}(2020)]%
        {chen2020context}
\bibfield{author}{\bibinfo{person}{Jia Chen}, \bibinfo{person}{Jiaxin Mao}, \bibinfo{person}{Yiqun Liu}, \bibinfo{person}{Min Zhang}, {and} \bibinfo{person}{Shaoping Ma}.} \bibinfo{year}{2020}\natexlab{}.
\newblock \showarticletitle{A context-aware click model for web search}. In \bibinfo{booktitle}{\emph{Proceedings of the 13th International Conference on Web Search and Data Mining}}. \bibinfo{pages}{88--96}.
\newblock


\bibitem[Chen et~al\mbox{.}(2021)]%
        {chen2021iobm}
\bibfield{author}{\bibinfo{person}{Mouxiang Chen}, \bibinfo{person}{Chenghao Liu}, \bibinfo{person}{Jianling Sun}, {and} \bibinfo{person}{Steven~CH Hoi}.} \bibinfo{year}{2021}\natexlab{}.
\newblock \showarticletitle{Adapting interactional observation embedding for counterfactual learning to rank}. In \bibinfo{booktitle}{\emph{Proceedings of the 44th International ACM SIGIR Conference on Research and Development in Information Retrieval}}. \bibinfo{pages}{285--294}.
\newblock


\bibitem[Craswell et~al\mbox{.}(2008)]%
        {craswell2008experimental}
\bibfield{author}{\bibinfo{person}{Nick Craswell}, \bibinfo{person}{Onno Zoeter}, \bibinfo{person}{Michael Taylor}, {and} \bibinfo{person}{Bill Ramsey}.} \bibinfo{year}{2008}\natexlab{}.
\newblock \showarticletitle{An experimental comparison of click position-bias models}. In \bibinfo{booktitle}{\emph{Proceedings of the 2008 international conference on web search and data mining}}. \bibinfo{pages}{87--94}.
\newblock


\bibitem[Dempster et~al\mbox{.}(1977)]%
        {dempster1977EM}
\bibfield{author}{\bibinfo{person}{Arthur~P Dempster}, \bibinfo{person}{Nan~M Laird}, {and} \bibinfo{person}{Donald~B Rubin}.} \bibinfo{year}{1977}\natexlab{}.
\newblock \showarticletitle{Maximum likelihood from incomplete data via the EM algorithm}.
\newblock \bibinfo{journal}{\emph{Journal of the royal statistical society: series B (methodological)}} \bibinfo{volume}{39}, \bibinfo{number}{1} (\bibinfo{year}{1977}), \bibinfo{pages}{1--22}.
\newblock


\bibitem[Dupret and Piwowarski(2008)]%
        {dupret2008ubm}
\bibfield{author}{\bibinfo{person}{Georges~E Dupret} {and} \bibinfo{person}{Benjamin Piwowarski}.} \bibinfo{year}{2008}\natexlab{}.
\newblock \showarticletitle{A user browsing model to predict search engine click data from past observations.}. In \bibinfo{booktitle}{\emph{Proceedings of the 31st annual international ACM SIGIR conference on Research and development in information retrieval}}. \bibinfo{pages}{331--338}.
\newblock


\bibitem[Guo et~al\mbox{.}(2009)]%
        {guo2009dcm}
\bibfield{author}{\bibinfo{person}{Fan Guo}, \bibinfo{person}{Chao Liu}, {and} \bibinfo{person}{Yi~Min Wang}.} \bibinfo{year}{2009}\natexlab{}.
\newblock \showarticletitle{Efficient multiple-click models in web search}. In \bibinfo{booktitle}{\emph{Proceedings of the second acm international conference on web search and data mining}}. \bibinfo{pages}{124--131}.
\newblock


\bibitem[Guo et~al\mbox{.}(2016)]%
        {guo2016deep}
\bibfield{author}{\bibinfo{person}{Jiafeng Guo}, \bibinfo{person}{Yixing Fan}, \bibinfo{person}{Qingyao Ai}, {and} \bibinfo{person}{W~Bruce Croft}.} \bibinfo{year}{2016}\natexlab{}.
\newblock \showarticletitle{A deep relevance matching model for ad-hoc retrieval}. In \bibinfo{booktitle}{\emph{Proceedings of the 25th ACM international on conference on information and knowledge management}}. \bibinfo{pages}{55--64}.
\newblock


\bibitem[Gupta et~al\mbox{.}(2023)]%
        {Gupta2023SafeCLTR}
\bibfield{author}{\bibinfo{person}{Shashank Gupta}, \bibinfo{person}{Harrie Oosterhuis}, {and} \bibinfo{person}{Maarten de Rijke}.} \bibinfo{year}{2023}\natexlab{}.
\newblock \showarticletitle{Safe Deployment for Counterfactual Learning to Rank with Exposure-Based Risk Minimization}. In \bibinfo{booktitle}{\emph{Proceedings of the 46th International ACM SIGIR Conference on Research and Development in Information Retrieval}} (Taipei, Taiwan) \emph{(\bibinfo{series}{SIGIR '23})}. \bibinfo{publisher}{Association for Computing Machinery}, \bibinfo{address}{New York, NY, USA}, \bibinfo{pages}{249–258}.
\newblock
\showISBNx{9781450394086}
\href{https://doi.org/10.1145/3539618.3591760}{doi:\nolinkurl{10.1145/3539618.3591760}}


\bibitem[Gupta et~al\mbox{.}(2024)]%
        {Gupta2024PRPO}
\bibfield{author}{\bibinfo{person}{Shashank Gupta}, \bibinfo{person}{Harrie Oosterhuis}, {and} \bibinfo{person}{Maarten de Rijke}.} \bibinfo{year}{2024}\natexlab{}.
\newblock \showarticletitle{Practical and Robust Safety Guarantees for Advanced Counterfactual Learning to Rank}. In \bibinfo{booktitle}{\emph{Proceedings of the 33rd ACM International Conference on Information and Knowledge Management}} (Boise, ID, USA) \emph{(\bibinfo{series}{CIKM '24})}. \bibinfo{publisher}{Association for Computing Machinery}, \bibinfo{address}{New York, NY, USA}, \bibinfo{pages}{737–747}.
\newblock
\showISBNx{9798400704369}
\href{https://doi.org/10.1145/3627673.3679531}{doi:\nolinkurl{10.1145/3627673.3679531}}


\bibitem[Hager et~al\mbox{.}(2024)]%
        {hager2024unbiased}
\bibfield{author}{\bibinfo{person}{Philipp Hager}, \bibinfo{person}{Romain Deffayet}, \bibinfo{person}{Jean-Michel Renders}, \bibinfo{person}{Onno Zoeter}, {and} \bibinfo{person}{Maarten de Rijke}.} \bibinfo{year}{2024}\natexlab{}.
\newblock \showarticletitle{Unbiased Learning to Rank Meets Reality: Lessons from Baidu's Large-Scale Search Dataset}. In \bibinfo{booktitle}{\emph{Proceedings of the 47th International ACM SIGIR Conference on Research and Development in Information Retrieval}}. \bibinfo{pages}{1546--1556}.
\newblock


\bibitem[J{\"a}rvelin and Kek{\"a}l{\"a}inen(2002)]%
        {jarvelin2002cumulated}
\bibfield{author}{\bibinfo{person}{Kalervo J{\"a}rvelin} {and} \bibinfo{person}{Jaana Kek{\"a}l{\"a}inen}.} \bibinfo{year}{2002}\natexlab{}.
\newblock \showarticletitle{Cumulated gain-based evaluation of IR techniques}.
\newblock \bibinfo{journal}{\emph{ACM Transactions on Information Systems (TOIS)}} \bibinfo{volume}{20}, \bibinfo{number}{4} (\bibinfo{year}{2002}), \bibinfo{pages}{422--446}.
\newblock


\bibitem[Joachims(2006)]%
        {joachims2006SVMRank}
\bibfield{author}{\bibinfo{person}{Thorsten Joachims}.} \bibinfo{year}{2006}\natexlab{}.
\newblock \showarticletitle{Training linear SVMs in linear time}. In \bibinfo{booktitle}{\emph{Proceedings of the 12th ACM SIGKDD international conference on Knowledge discovery and data mining}}. \bibinfo{pages}{217--226}.
\newblock


\bibitem[Joachims et~al\mbox{.}(2005)]%
        {joachims2005accurately}
\bibfield{author}{\bibinfo{person}{Thorsten Joachims}, \bibinfo{person}{Laura Granka}, \bibinfo{person}{Bing Pan}, \bibinfo{person}{Helene Hembrooke}, {and} \bibinfo{person}{Geri Gay}.} \bibinfo{year}{2005}\natexlab{}.
\newblock \showarticletitle{Accurately Interpreting Clickthrough Data as Implicit Feedback}.
\newblock  (\bibinfo{year}{2005}).
\newblock


\bibitem[Joachims et~al\mbox{.}(2007)]%
        {joachims2007evaluating}
\bibfield{author}{\bibinfo{person}{Thorsten Joachims}, \bibinfo{person}{Laura Granka}, \bibinfo{person}{Bing Pan}, \bibinfo{person}{Helene Hembrooke}, \bibinfo{person}{Filip Radlinski}, {and} \bibinfo{person}{Geri Gay}.} \bibinfo{year}{2007}\natexlab{}.
\newblock \showarticletitle{Evaluating the accuracy of implicit feedback from clicks and query reformulations in web search}.
\newblock \bibinfo{journal}{\emph{ACM Transactions on Information Systems (TOIS)}} \bibinfo{volume}{25}, \bibinfo{number}{2} (\bibinfo{year}{2007}), \bibinfo{pages}{7--es}.
\newblock


\bibitem[Joachims et~al\mbox{.}(2017)]%
        {joachims2017ips}
\bibfield{author}{\bibinfo{person}{Thorsten Joachims}, \bibinfo{person}{Adith Swaminathan}, {and} \bibinfo{person}{Tobias Schnabel}.} \bibinfo{year}{2017}\natexlab{}.
\newblock \showarticletitle{Unbiased learning-to-rank with biased feedback}. In \bibinfo{booktitle}{\emph{Proceedings of the Tenth ACM International Conference on Web Search and Data Mining}}. \bibinfo{pages}{781--789}.
\newblock


\bibitem[Luo et~al\mbox{.}(2024)]%
        {Luo2024UPE}
\bibfield{author}{\bibinfo{person}{Dan Luo}, \bibinfo{person}{Lixin Zou}, \bibinfo{person}{Qingyao Ai}, \bibinfo{person}{Zhiyu Chen}, \bibinfo{person}{Chenliang Li}, \bibinfo{person}{Dawei Yin}, {and} \bibinfo{person}{Brian~D. Davison}.} \bibinfo{year}{2024}\natexlab{}.
\newblock \showarticletitle{Unbiased Learning-to-Rank Needs Unconfounded Propensity Estimation}. In \bibinfo{booktitle}{\emph{Proceedings of the 47th International ACM SIGIR Conference on Research and Development in Information Retrieval}} (Washington DC, USA) \emph{(\bibinfo{series}{SIGIR '24})}. \bibinfo{publisher}{Association for Computing Machinery}, \bibinfo{address}{New York, NY, USA}, \bibinfo{pages}{1535–1545}.
\newblock
\showISBNx{9798400704314}
\href{https://doi.org/10.1145/3626772.3657772}{doi:\nolinkurl{10.1145/3626772.3657772}}


\bibitem[Mao et~al\mbox{.}(2018)]%
        {mao2018MCM}
\bibfield{author}{\bibinfo{person}{Jiaxin Mao}, \bibinfo{person}{Cheng Luo}, \bibinfo{person}{Min Zhang}, {and} \bibinfo{person}{Shaoping Ma}.} \bibinfo{year}{2018}\natexlab{}.
\newblock \showarticletitle{Constructing click models for mobile search}. In \bibinfo{booktitle}{\emph{The 41st International ACM SIGIR Conference on Research \& Development in Information Retrieval}}. \bibinfo{pages}{775--784}.
\newblock


\bibitem[Mitra et~al\mbox{.}(2017)]%
        {mitra2017learning}
\bibfield{author}{\bibinfo{person}{Bhaskar Mitra}, \bibinfo{person}{Fernando Diaz}, {and} \bibinfo{person}{Nick Craswell}.} \bibinfo{year}{2017}\natexlab{}.
\newblock \showarticletitle{Learning to match using local and distributed representations of text for web search}. In \bibinfo{booktitle}{\emph{Proceedings of the 26th international conference on world wide web}}. \bibinfo{pages}{1291--1299}.
\newblock


\bibitem[Oosterhuis and de~Rijke(2018)]%
        {oosterhuis2018PDGD}
\bibfield{author}{\bibinfo{person}{Harrie Oosterhuis} {and} \bibinfo{person}{Maarten de Rijke}.} \bibinfo{year}{2018}\natexlab{}.
\newblock \showarticletitle{Differentiable unbiased online learning to rank}. In \bibinfo{booktitle}{\emph{Proceedings of the 27th ACM international conference on information and knowledge management}}. \bibinfo{pages}{1293--1302}.
\newblock


\bibitem[Oosterhuis and de~Rijke(2020)]%
        {oosterhuis2020policy-aware}
\bibfield{author}{\bibinfo{person}{Harrie Oosterhuis} {and} \bibinfo{person}{Maarten de Rijke}.} \bibinfo{year}{2020}\natexlab{}.
\newblock \showarticletitle{Policy-aware unbiased learning to rank for top-k rankings}. In \bibinfo{booktitle}{\emph{Proceedings of the 43rd International ACM SIGIR Conference on Research and Development in Information Retrieval}}. \bibinfo{pages}{489--498}.
\newblock


\bibitem[Oosterhuis and de~Rijke(2022)]%
        {oosterhuis2022reaching}
\bibfield{author}{\bibinfo{person}{Harrie Oosterhuis} {and} \bibinfo{person}{Maarten de Rijke}.} \bibinfo{year}{2022}\natexlab{}.
\newblock \showarticletitle{Reaching the End of Unbiasedness: Uncovering Implicit Limitations of Click-Based Learning to Rank}. In \bibinfo{booktitle}{\emph{Proceedings of the 2022 ACM SIGIR International Conference on the Theory of Information Retrieval. ACM}}.
\newblock


\bibitem[Ovaisi et~al\mbox{.}(2020)]%
        {ovaisi2020selectionbias}
\bibfield{author}{\bibinfo{person}{Zohreh Ovaisi}, \bibinfo{person}{Ragib Ahsan}, \bibinfo{person}{Yifan Zhang}, \bibinfo{person}{Kathryn Vasilaky}, {and} \bibinfo{person}{Elena Zheleva}.} \bibinfo{year}{2020}\natexlab{}.
\newblock \showarticletitle{Correcting for selection bias in learning-to-rank systems}. In \bibinfo{booktitle}{\emph{Proceedings of The Web Conference 2020}}. \bibinfo{pages}{1863--1873}.
\newblock


\bibitem[O’Brien and Keane(2006)]%
        {o2006modeling}
\bibfield{author}{\bibinfo{person}{Maeve O’Brien} {and} \bibinfo{person}{Mark~T Keane}.} \bibinfo{year}{2006}\natexlab{}.
\newblock \showarticletitle{Modeling result-list searching in the World Wide Web: The role of relevance topologies and trust bias}. In \bibinfo{booktitle}{\emph{Proceedings of the 28th annual conference of the cognitive science society}}, Vol.~\bibinfo{volume}{28}. Citeseer, \bibinfo{pages}{1881--1886}.
\newblock


\bibitem[Qin and Liu(2013)]%
        {MSLR}
\bibfield{author}{\bibinfo{person}{Tao Qin} {and} \bibinfo{person}{Tie{-}Yan Liu}.} \bibinfo{year}{2013}\natexlab{}.
\newblock \showarticletitle{Introducing {LETOR} 4.0 Datasets}.
\newblock \bibinfo{journal}{\emph{CoRR}}  \bibinfo{volume}{abs/1306.2597} (\bibinfo{year}{2013}).
\newblock
\urldef\tempurl%
\url{http://arxiv.org/abs/1306.2597}
\showURL{%
\tempurl}


\bibitem[Richardson et~al\mbox{.}(2007)]%
        {richardson2007ExminationHypothesis}
\bibfield{author}{\bibinfo{person}{Matthew Richardson}, \bibinfo{person}{Ewa Dominowska}, {and} \bibinfo{person}{Robert Ragno}.} \bibinfo{year}{2007}\natexlab{}.
\newblock \showarticletitle{Predicting clicks: estimating the click-through rate for new ads}. In \bibinfo{booktitle}{\emph{Proceedings of the 16th international conference on World Wide Web}}. \bibinfo{pages}{521--530}.
\newblock


\bibitem[Rosenbaum and Rubin(1983)]%
        {ipw}
\bibfield{author}{\bibinfo{person}{Paul~R Rosenbaum} {and} \bibinfo{person}{Donald~B Rubin}.} \bibinfo{year}{1983}\natexlab{}.
\newblock \showarticletitle{The central role of the propensity score in observational studies for causal effects}.
\newblock \bibinfo{journal}{\emph{Biometrika}} \bibinfo{volume}{70}, \bibinfo{number}{1} (\bibinfo{year}{1983}), \bibinfo{pages}{41--55}.
\newblock


\bibitem[Schuth et~al\mbox{.}(2016)]%
        {schuth2016MGD}
\bibfield{author}{\bibinfo{person}{Anne Schuth}, \bibinfo{person}{Harrie Oosterhuis}, \bibinfo{person}{Shimon Whiteson}, {and} \bibinfo{person}{Maarten de Rijke}.} \bibinfo{year}{2016}\natexlab{}.
\newblock \showarticletitle{Multileave gradient descent for fast online learning to rank}. In \bibinfo{booktitle}{\emph{proceedings of the ninth ACM international conference on web search and data mining}}. \bibinfo{pages}{457--466}.
\newblock


\bibitem[Schuth et~al\mbox{.}(2014)]%
        {schuth2014multileaved}
\bibfield{author}{\bibinfo{person}{Anne Schuth}, \bibinfo{person}{Floor Sietsma}, \bibinfo{person}{Shimon Whiteson}, \bibinfo{person}{Damien Lefortier}, {and} \bibinfo{person}{Maarten de Rijke}.} \bibinfo{year}{2014}\natexlab{}.
\newblock \showarticletitle{Multileaved comparisons for fast online evaluation}. In \bibinfo{booktitle}{\emph{Proceedings of the 23rd ACM International Conference on Conference on Information and Knowledge Management}}. \bibinfo{pages}{71--80}.
\newblock


\bibitem[Vardasbi et~al\mbox{.}(2020a)]%
        {vardasbi2020cm-ips}
\bibfield{author}{\bibinfo{person}{Ali Vardasbi}, \bibinfo{person}{Maarten de Rijke}, {and} \bibinfo{person}{Ilya Markov}.} \bibinfo{year}{2020}\natexlab{a}.
\newblock \showarticletitle{Cascade model-based propensity estimation for counterfactual learning to rank}. In \bibinfo{booktitle}{\emph{Proceedings of the 43rd International ACM SIGIR Conference on Research and Development in Information Retrieval}}. \bibinfo{pages}{2089--2092}.
\newblock


\bibitem[Vardasbi et~al\mbox{.}(2020b)]%
        {vardasbi2020affine}
\bibfield{author}{\bibinfo{person}{Ali Vardasbi}, \bibinfo{person}{Harrie Oosterhuis}, {and} \bibinfo{person}{Maarten de Rijke}.} \bibinfo{year}{2020}\natexlab{b}.
\newblock \showarticletitle{When inverse propensity scoring does not work: Affine corrections for unbiased learning to rank}. In \bibinfo{booktitle}{\emph{Proceedings of the 29th ACM International Conference on Information \& Knowledge Management}}. \bibinfo{pages}{1475--1484}.
\newblock


\bibitem[Wang et~al\mbox{.}(2019)]%
        {wang2019variance}
\bibfield{author}{\bibinfo{person}{Huazheng Wang}, \bibinfo{person}{Sonwoo Kim}, \bibinfo{person}{Eric McCord-Snook}, \bibinfo{person}{Qingyun Wu}, {and} \bibinfo{person}{Hongning Wang}.} \bibinfo{year}{2019}\natexlab{}.
\newblock \showarticletitle{Variance reduction in gradient exploration for online learning to rank}. In \bibinfo{booktitle}{\emph{Proceedings of the 42nd International ACM SIGIR Conference on Research and Development in Information Retrieval}}. \bibinfo{pages}{835--844}.
\newblock


\bibitem[Wang et~al\mbox{.}(2018b)]%
        {wang2018efficient}
\bibfield{author}{\bibinfo{person}{Huazheng Wang}, \bibinfo{person}{Ramsey Langley}, \bibinfo{person}{Sonwoo Kim}, \bibinfo{person}{Eric McCord-Snook}, {and} \bibinfo{person}{Hongning Wang}.} \bibinfo{year}{2018}\natexlab{b}.
\newblock \showarticletitle{Efficient exploration of gradient space for online learning to rank}. In \bibinfo{booktitle}{\emph{The 41st international ACM SIGIR conference on research \& development in information retrieval}}. \bibinfo{pages}{145--154}.
\newblock


\bibitem[Wang et~al\mbox{.}(2021)]%
        {wang2021prs}
\bibfield{author}{\bibinfo{person}{Nan Wang}, \bibinfo{person}{Zhen Qin}, \bibinfo{person}{Xuanhui Wang}, {and} \bibinfo{person}{Hongning Wang}.} \bibinfo{year}{2021}\natexlab{}.
\newblock \showarticletitle{Non-clicks mean irrelevant? propensity ratio scoring as a correction}. In \bibinfo{booktitle}{\emph{Proceedings of the 14th ACM International Conference on Web Search and Data Mining}}. \bibinfo{pages}{481--489}.
\newblock


\bibitem[Wang et~al\mbox{.}(2016)]%
        {wang2016ips}
\bibfield{author}{\bibinfo{person}{Xuanhui Wang}, \bibinfo{person}{Michael Bendersky}, \bibinfo{person}{Donald Metzler}, {and} \bibinfo{person}{Marc Najork}.} \bibinfo{year}{2016}\natexlab{}.
\newblock \showarticletitle{Learning to rank with selection bias in personal search}. In \bibinfo{booktitle}{\emph{Proceedings of the 39th International ACM SIGIR conference on Research and Development in Information Retrieval}}. \bibinfo{pages}{115--124}.
\newblock


\bibitem[Wang et~al\mbox{.}(2018a)]%
        {wang2018regression-EM}
\bibfield{author}{\bibinfo{person}{Xuanhui Wang}, \bibinfo{person}{Nadav Golbandi}, \bibinfo{person}{Michael Bendersky}, \bibinfo{person}{Donald Metzler}, {and} \bibinfo{person}{Marc Najork}.} \bibinfo{year}{2018}\natexlab{a}.
\newblock \showarticletitle{Position bias estimation for unbiased learning to rank in personal search}. In \bibinfo{booktitle}{\emph{Proceedings of the Eleventh ACM International Conference on Web Search and Data Mining}}. \bibinfo{pages}{610--618}.
\newblock


\bibitem[Yue and Joachims(2009)]%
        {yue2009DBGD}
\bibfield{author}{\bibinfo{person}{Yisong Yue} {and} \bibinfo{person}{Thorsten Joachims}.} \bibinfo{year}{2009}\natexlab{}.
\newblock \showarticletitle{Interactively optimizing information retrieval systems as a dueling bandits problem}. In \bibinfo{booktitle}{\emph{Proceedings of the 26th Annual International Conference on Machine Learning}}. \bibinfo{pages}{1201--1208}.
\newblock


\bibitem[Yue et~al\mbox{.}(2010)]%
        {yue2010beyond}
\bibfield{author}{\bibinfo{person}{Yisong Yue}, \bibinfo{person}{Rajan Patel}, {and} \bibinfo{person}{Hein Roehrig}.} \bibinfo{year}{2010}\natexlab{}.
\newblock \showarticletitle{Beyond position bias: Examining result attractiveness as a source of presentation bias in clickthrough data}. In \bibinfo{booktitle}{\emph{Proceedings of the 19th international conference on World wide web}}. \bibinfo{pages}{1011--1018}.
\newblock


\bibitem[Zhang et~al\mbox{.}(2021)]%
        {zhang2021cbcm}
\bibfield{author}{\bibinfo{person}{Ruizhe Zhang}, \bibinfo{person}{Xiaohui Xie}, \bibinfo{person}{Jiaxin Mao}, \bibinfo{person}{Yiqun Liu}, \bibinfo{person}{Min Zhang}, {and} \bibinfo{person}{Shaoping Ma}.} \bibinfo{year}{2021}\natexlab{}.
\newblock \showarticletitle{Constructing a comparison-based click model for web search}. In \bibinfo{booktitle}{\emph{Proceedings of the Web Conference 2021}}. \bibinfo{pages}{270--283}.
\newblock


\bibitem[Zhao and King(2016)]%
        {zhao2016constructing}
\bibfield{author}{\bibinfo{person}{Tong Zhao} {and} \bibinfo{person}{Irwin King}.} \bibinfo{year}{2016}\natexlab{}.
\newblock \showarticletitle{Constructing reliable gradient exploration for online learning to rank}. In \bibinfo{booktitle}{\emph{Proceedings of the 25th ACM International on Conference on Information and Knowledge Management}}. \bibinfo{pages}{1643--1652}.
\newblock


\bibitem[Zou et~al\mbox{.}({[n.\,d.]})]%
        {Baidu}
\bibfield{author}{\bibinfo{person}{Lixin Zou}, \bibinfo{person}{Haitao Mao}, \bibinfo{person}{Xiaokai Chu}, \bibinfo{person}{Jiliang Tang}, \bibinfo{person}{Wenwen Ye}, \bibinfo{person}{Shuaiqiang Wang}, {and} \bibinfo{person}{Dawei Yin}.} \bibinfo{year}{[n.\,d.]}\natexlab{}.
\newblock \showarticletitle{A Large Scale Search Dataset for Unbiased Learning to Rank}. In \bibinfo{booktitle}{\emph{Thirty-sixth Conference on Neural Information Processing Systems Datasets and Benchmarks Track}}.
\newblock


\end{thebibliography}

\end{document}